\documentclass[letterpaper, 10 pt, conference]{ieeeconf}  

\IEEEoverridecommandlockouts                              

\usepackage{graphics} 
\usepackage{epsfig} 
\usepackage{mathptmx} 
\usepackage{times} 
\usepackage{amsmath} 
\usepackage{amssymb}  
\usepackage{graphicx}
\usepackage{subfigure}
\usepackage{stfloats}
\usepackage{url}
\usepackage{hyperref }
\usepackage{array}

\title{\LARGE \bf
Deep Reinforcement Learning for Robotic Pushing and Picking ~~~~~~~~~~~~~~~~in Cluttered Environment}

\author{Yuhong Deng$^{*}$, Xiaofeng Guo$^{*}$, Yixuan Wei$^{*}$, Kai Lu$^{*}$, Bin Fang, Di Guo, Huaping Liu$^{\dagger}$, Fuchun Sun
\thanks{* denotes equal contribution.}
\thanks{$\dagger$ Corresponding author: \tt\small hpliu@tsinghua.edu.cn}
\thanks{All authors are from Department of Computer Science and Technology, Tsinghua University, Beijing, China. Di Guo is also with Shenzhen Academy of Robotics.}
}

\begin{document}

\maketitle
\thispagestyle{empty}
\pagestyle{empty}


\begin{abstract}

In this paper, a novel robotic grasping system is established to automatically pick up objects in cluttered scenes. A composite robotic hand composed of a suction cup and a gripper is designed for grasping the object stably. The suction cup is used for lifting the object from the clutter first and the gripper for grasping the object accordingly. We utilize the affordance map to provide pixel-wise lifting point candidates for the suction cup. To obtain a good affordance map, the active exploration mechanism is introduced to the system. An effective metric is designed to calculate the reward for the current affordance map, and a deep Q-Network (DQN) is employed to guide the robotic hand to actively explore the environment until the generated affordance map is suitable for grasping. Experimental results have demonstrated that the proposed robotic grasping system is able to greatly increase the success rate of the robotic grasping in cluttered scenes.

\end{abstract}

\section{INTRODUCTION}

With the rapid development of e-commerce, a growing demand has been put on using autonomous robots in logistics. There have been already a lot of mobile robots working at real warehouses for product transportation. It is still a great challenge for the robot to pick and sort products in real scenarios automatically \cite{liu2017material}. This kind of work is highly dependent on human workers nowadays, which is not economically and time efficient. In this work, we propose a novel robotic grasping system which is able to automatically pick up objects in cluttered scenes. A composite robotic hand which is able to grasp many kinds of different objects robustly is designed. A deep Q-Network (DQN) \cite{DQN} is employed to guide the robotic hand to actively explore the environment to find proper grasping points.

The design of robotic hands have been studied for years, and many different types of robotic hands have been proposed. The robotic hand with suction cup is very popular and widely used in robotic grasping tasks \cite{suck_soft, Dex}.  It is because that the suction cup is usually with a simple structure and robust to many different objects. In \cite{suction_cup_arrays}, self-sealing suction cup arrays are proposed to greatly improve the robotic grasping ability in uncertain environments. To increase the adhesion force of the suction cup, a stretchable suction cup with electroadhesion is designed \cite{suction_cup_ele}. There are also some other suction cups inspired by biometric designs \cite{suction_cup_cep,suction_cup_octopus,suction_cup_urchin,suction_cup_oct2,kuwajima2017active}. However, the working mechanism of the suction cup imposes many restrictions on the surface and postures of the object. In addition, the inconsistency between the moving direction of the suction cup and the force direction makes the grasping unstable \cite{mantriota2011theoretical} and causes the working life of the suction cup short. Therefore, it is important for the robotic hand to find proper grasping points when using the suction cup.

\begin{figure}
\centering
\includegraphics[width=1\linewidth]{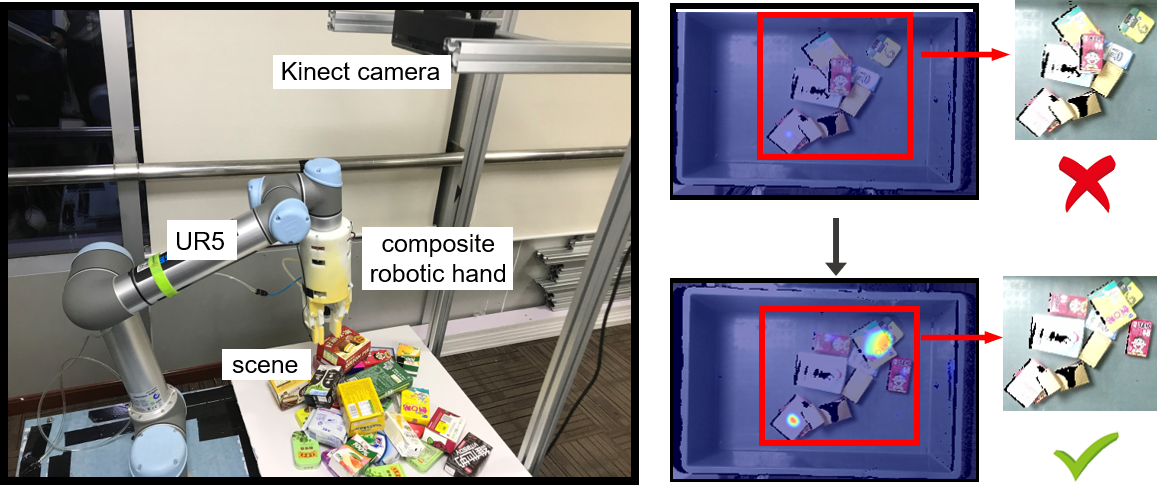}
\caption{\textbf{Grasping System.} Our grasping system consists of a composite robotic hand, a UR5 manipulator, and a Kinect camera. An active exploration strategy is integrated to this system to implement effective robotic grasping.}
\label{fig:concept}
\end{figure}

Zeng et al. \cite{affordance} have proposed to use affordance map to indicate grasping points by analyzing the whole scene, which greatly improves the accuracy of robotic grasping. The affordance map is a graph which shows the confidence rate of each pixel in the input image for grasping. However, since the environment is usually complex and unstructured, sometimes the grasping location that the affordance map indicates is difficult for the robot to grasp. To solve this problem, the active exploration mechanism is introduced \cite{active_ex_perception,active_vision,liu2018robotic}. By actively exploring the environment, the robot is able to make some changes to the environment until it is suitable for grasping. For example, when the objects in the scene are too close to each other for grasping, the robot can actively explore the environment and change the position of the object until it is suitable for grasping. Similarly, the robot can rearrange positions of objects by pushing them apart \cite{pushactive}.

In this paper, a composite robotic hand composed of a suction cup and a gripper is designed. With a deep Q-Network (DQN), the robotic hand can actively explore the environment until a good affordance map is obtained. The whole grasp system (Fig. \ref{fig:concept}) is able to effectively grasp many kinds of objects in a real cluttered environment. The main contributions are summarized as follows:

\begin{itemize}
  \item A novel composite robotic hand which combines a suction cup and a gripper is designed. It is able to grasp different objects quickly and stably.
  \item An active exploration algorithm which leverages the deep Q-Network (DQN) is proposed to facilitate the robot to actively explore the environment until a good affordance map is generated.
  \item The composite hand and the active exploration algorithm are fully integrated and the experimental results demonstrate the superior performance of this system when grasping objects in a real cluttered environment.

\end{itemize}

The rest of this paper is organized as follows. Some related work is introduced in Section \ref{sec:Related work}. A brief overview of the proposed robotic grasping system is presented in Section \ref{sec:System overview}. Section \ref{sec:Architecture} and Section \ref{sec:Active exploration} describes the composite robotic hand we designed and the grasping strategy in details. Extensive experimental results are demonstrated in Section \ref{sec:Experiment} to verify the effectiveness of the proposed robotic grasping system.

\section{RELATED WORK}
\label{sec:Related work}
\label{sec:System overview}

Robotic grasping is playing a more and more important role in many application areas such as industrial manufacturing, domestic service, logistics, etc. Because of the diversity and complexity of both the object and environment, higher requirements have been placed on the design of robotic hands. In general, robotic hands can be divided into two classes: 1) robotic hand with suction cup and 2) multi-finger robotic hand. Either design has its own specific application scenario. It is difficult to use only one single operation mode to fulfill all the tasks.

Therefore, many researchers try to leverage advantages of both types of robotic hands and some composite robotic hands are proposed \cite{Kessens2016Versatile,three_fingered_hand,stuart2015suction}. For example, a multi-functional gripper with a retractable mechanism is designed, which can switch between suction mode and grasping mode quickly and automatically \cite{affordance}. It provides a hardware basis for implementing different grasp strategies. However, this multi-functional gripper doesn't consider the coupling between the two modes. It can only choose to execute one operation mode at a time. In addition, Hasegawa et al. \cite{three_fingered_hand} propose the Suction Pinching Hand, which has two underactuated fingers and one extendable and foldable finger whose fingertip has a suction cup mounted on it. Experiments have shown that it can grasp various objects stably by using both the suction and pinching at the same time. Some other similar type of robotic hand has already been used in industrial solutions\cite{righthand}, \cite{cobot}.

Compared with the above composite robotic hand, the robotic hand proposed in this paper is of a much more simple and flexible structure. It seamlessly combines both a suction cup and a two-finger gripper. It has a suction mode and a grasping mode, which can be coupled to work simultaneously and also work separately. What's more, the proposed composite robotic hand is able to close its two fingers to push objects in order to actively explore the environment. A preliminary version of this paper has been published in \cite{active}, which discussed this robotic finger but it did not provide the design guidelines of the reward. In this paper, we present more detailed illustrations about the experimental results and provide the details about the reward function design for the deep reinforcement learning.

\section{SYSTEM OVERVIEW}
\begin{figure*}[ht]
\centering
\includegraphics[width=16cm]{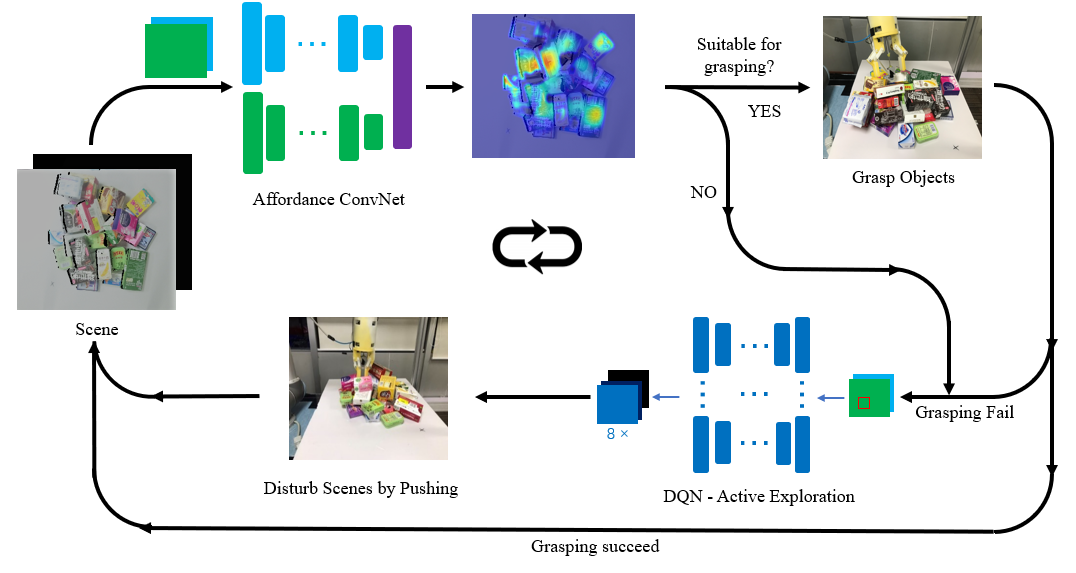}
\caption{\textbf{System Pipeline.} The system firstly obtains the RGB image and depth image of the scene, and the images are fed into the affordance ConvNet to calculate the affordance map. The system will then evaluate the quality of the affordance map by a metric. If the affordance map is not good enough, the current affordance map will be fed into the DQN, which can suggest an appropriate operation for the robotic hand to explore the environment. Otherwise, the composite robotic hand will implement the suction operation and then grasp the object.}
\label{fig:system}
\end{figure*}

The pipeline of the proposed robotic grasping system is illustrated in Fig. \ref{fig:system}. The RGB image and depth image of the scene are obtained firstly. The affordance ConvNet \cite{ affordance} is then used to calculate the affordance map based on the input images. A metric $\Phi$ is proposed to evaluate the quality of the current affordance map. If $\Phi$ is above a certain threshold value, the composite robotic hand will implement the suction operation with the suction cup and then grasp the object accordingly. Otherwise, the affordance map will be fed into the DQN, which guides the composite robotic hand to make some disturbances to the environment by pushing objects in the front. This process will be iterated until all the objects in the environment are successfully picked up.  

\section{ARCHITECTURE}
\label{sec:Architecture}

At present, the robotic hands can be mainly divided into two classes: 1) robotic hand with suction cup and 2) multi-finger robotic hand. Either design of the robotic hand has its own characteristics. It is reasonable to leverage advantages of both robotic hands to construct a composite one. Therefore, we propose a new type of composite robotic hand, which combines a gripper and a suction cup together.

\subsection{Robotic Hand Structure}
\begin{figure}[!hb]
\centering
\includegraphics[width=\linewidth]{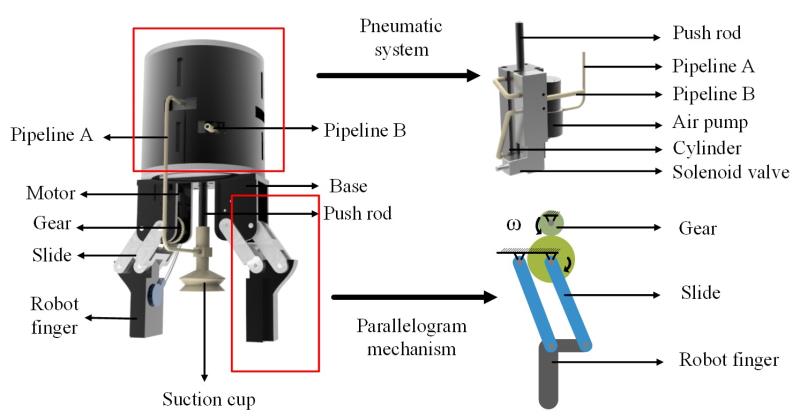}
\caption{\textbf{Structure of the Composite Robotic Hand.} The composite robotic hand is composed of two parallel fingers and a suction cup system. }
\label{fig:hand_structure}
\end{figure}

The structure of our composite robotic hand is shown in Fig. \ref{fig:hand_structure}. The composite robotic hand is composed of two parallel fingers and a suction cup. The two fingers are symmetrically distributed on the base. There is a motor-driven parallelogram mechanism for each finger, which ensures that the surfaces of the two fingers are always parallel when the finger is grasping the object.

The suction cup system consists of a suction cup, a push rod, a cylinder, two air pumps, a miniature motor and a solenoid valve. The suction cup is placed in the middle of the two fingers. Two air pumps are respecptively equipped inside and outside of the composite robotic hand. The inside one and miniature motor are used for controlling the suction cup, while the outside one with solenoid valve drives the push rod with a range of 75mm.

\label{sec:System overview}
\subsection{Grasp Process}

\begin{figure}[h]
\centering
\subfigure[localization]{
\begin{minipage}[t]{0.32\linewidth}
\centering
\includegraphics[width=1in]{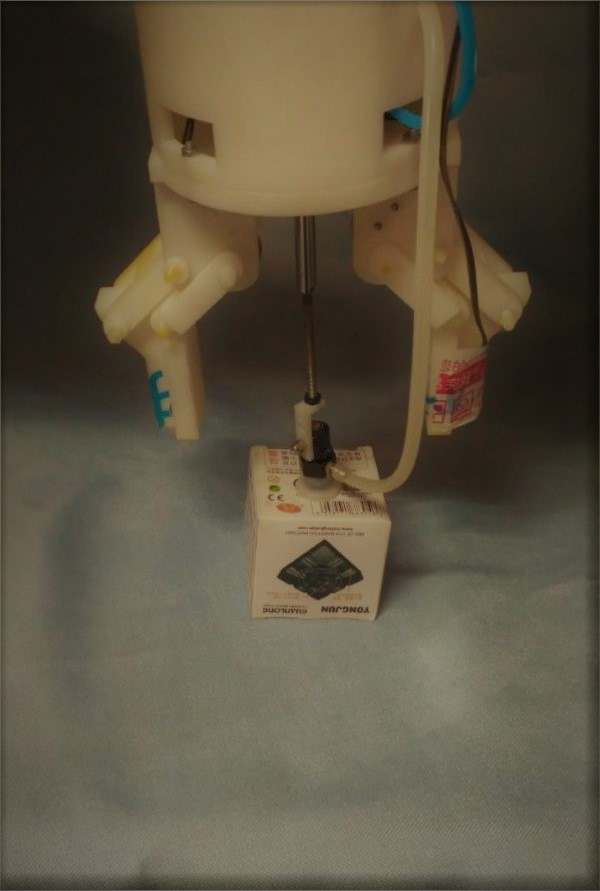}
\end{minipage}%
}%
\subfigure[suction]{
\begin{minipage}[t]{0.32\linewidth}
\centering
\includegraphics[width=1in]{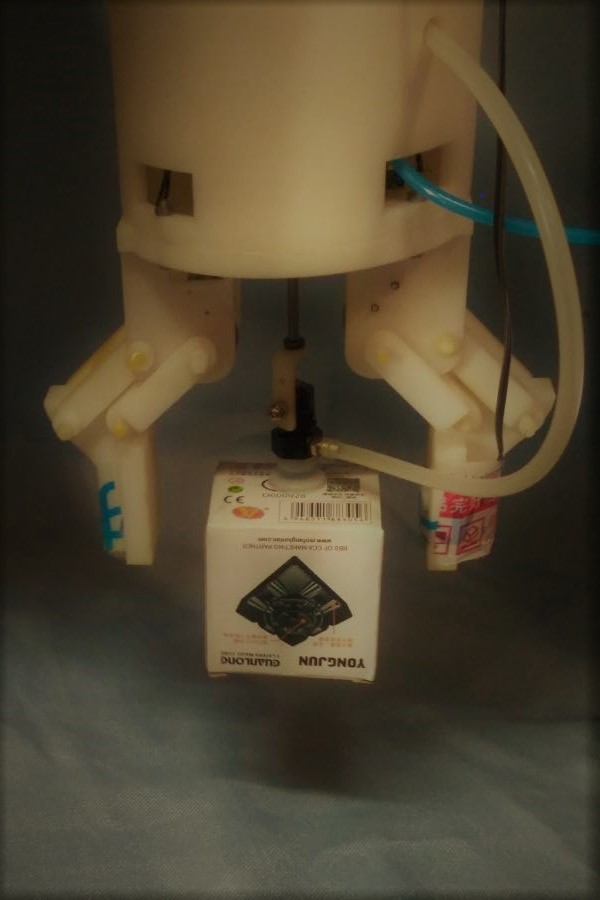}
\end{minipage}%
}%
\subfigure[grasping]{
\begin{minipage}[t]{0.32\linewidth}
\centering
\includegraphics[width=1in]{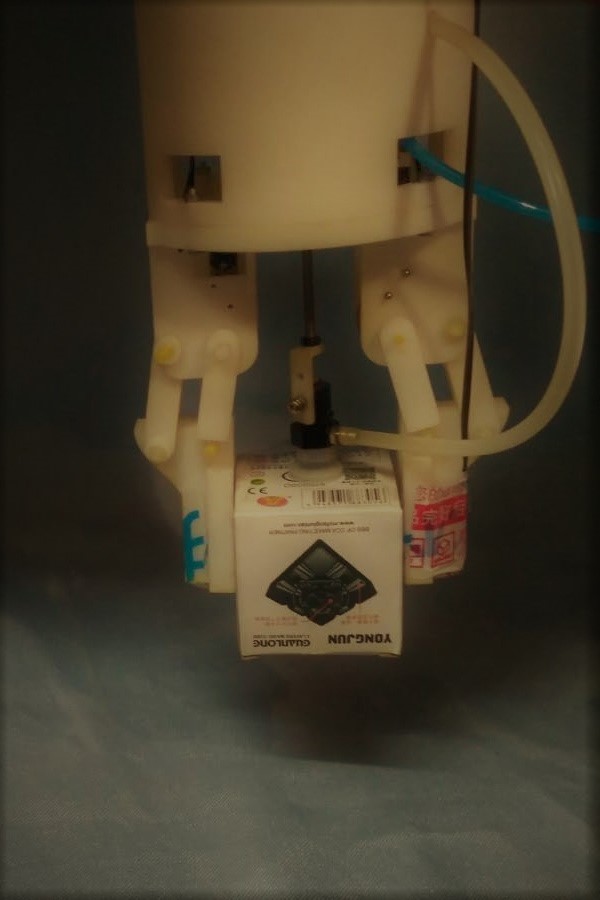}
\end{minipage}
}%
\centering
\caption{\textbf{Object Grasping Process.} The suction cup extends first to lift the object, and then the gripper closes to grasp the object.
 }
\label{fig:grasp_process}
\end{figure}

During the process of grasping (Fig. \ref{fig:grasp_process}), the two fingers are in an open state, and the suction cup is held in the initial state. The robotic hand moves to the lifting point, and when the lifting point is reached, the suction cup will be popped out to approach the surface of the objects. Then the air pump generates the negative pressure in the suction cup so that the object is lifted. Next, the push rod retracts to take the object between the two fingers. Finally, the fingers close to ensure the stability of the grasp. At last, the object will be released. The process of releasing the object is opposite to the suction process.

\subsection{Characteristics of Grasp Process}
\begin{figure}[!h]
\centering
\includegraphics[width=1\linewidth]{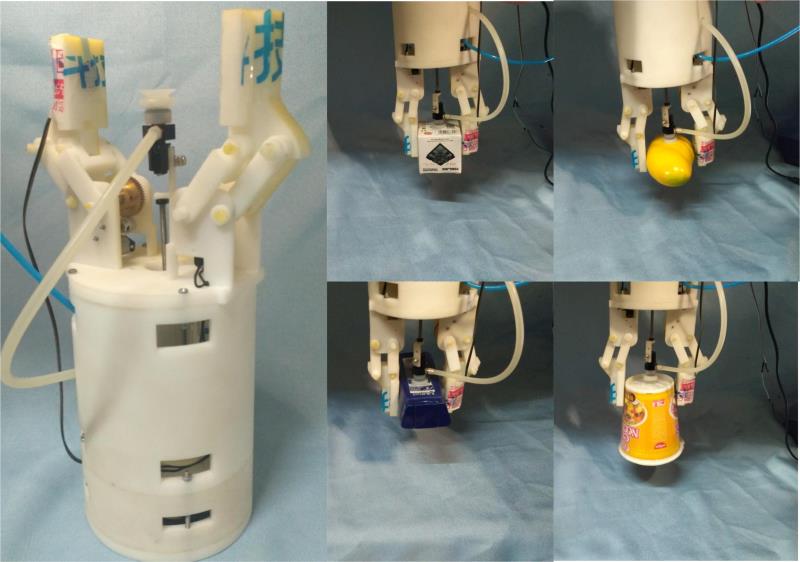}
\caption{\textbf{Prototype and Grasp Experiment.} The composite robotic hand is grasping several objects of different shapes.}
\label{fig:hand_protype}
\end{figure}

Compared with other suction grasping systems, the proposed composite robotic hand uses the two fingers to hold the object after the suction cup lifts the object, which increases the stability of the grasp. Especially, when the robotic hand is moving, the force applied by the fingers and suction cup can coordinate together to guarantee the object is stably grasped. Experiments have proved that our composite robotic hand can grasp objects of different sizes and shapes effectively and stably. Some results are demonstrated in Fig. \ref{fig:hand_protype}.

\section{Deep Q-Network Structure}
\label{sec:Active exploration}
\begin{figure*}[ht]
	\centering
	\includegraphics[width=\textwidth]{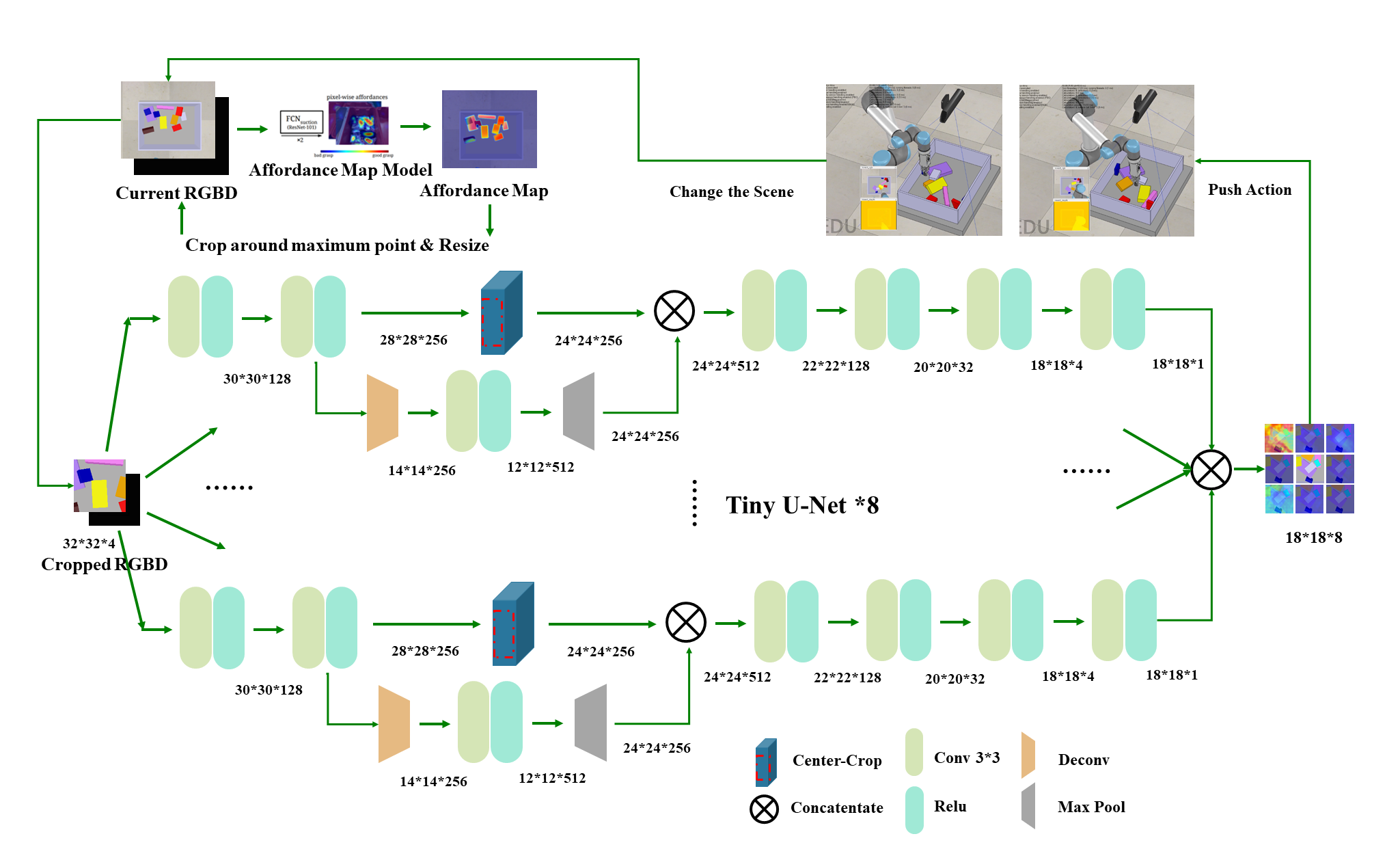}
	\caption{\textbf{Deep Q-Network Structure.} For current RGBD frame, we utilize the affordance map to output a primitive operation guidance, and crop and resize the input frame around the pixel with maximum confidence. We feed this local patch into 8 paralleled tiny U-Net and 8 diverse action directions are output on subpixel-wise locations. The reward for this network is calculated according to a specific designed metric which is derived from the affordance map.}
	\label{fig:DQN_Stru}
\end{figure*}
\subsection{Affordance Map}
\label{sec:Affordance map}
The affordance map is a graph which shows the confidence rate of each point in input images for grasping \cite{affordance}. In this paper, it is used to provide pixel-wise lifting point candidates for the suction cup. It solves the problem in traditional grasping strategy that requires to recognize the object first before grasping. However, it is inevitable that sometimes it is hard to distinguish good grasping points from the obtained affordance map, especially when the scenario is complicated. In this situation, we propose that the robot is supposed to have the ability to actively explore and change the environment until a good affordance map is obtained.
\par
\subsubsection{Affordance ConvNet}
Affordance ConvNet \cite{affordance} is a network which takes RGB and depth images as input and outputs the affordance map which is a dense pixel-wise heatmap with values ranging from 0 to 1. The closer the values are to 1, the more preferable the lifting locations are. For training purpose, we manually label the scene images, in which areas that are suitable for grasping are annotated.

\subsubsection{Failure cases}
\begin{figure}[h]
\centering
\subfigure[gathering]{
\begin{minipage}[t]{0.3\linewidth}
\centering
\includegraphics[width=0.9in]{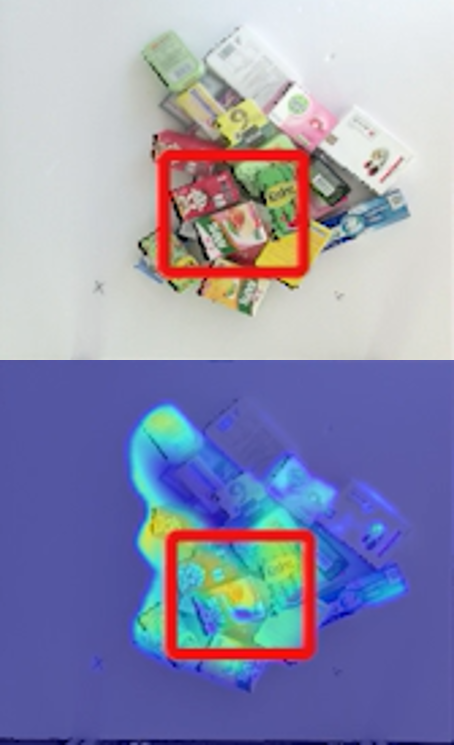}
\label{fig:gather}
\end{minipage}
}%
\subfigure[covering]{
\begin{minipage}[t]{0.3\linewidth}
\centering
\includegraphics[width=0.9in]{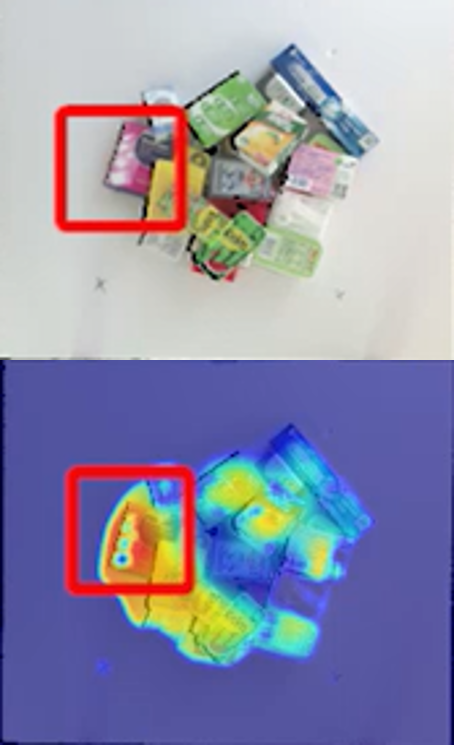}
\end{minipage}
\label{fig:cover}
}%
\subfigure[tilting]{
\begin{minipage}[t]{0.3\linewidth}
\centering
\includegraphics[width=0.9in]{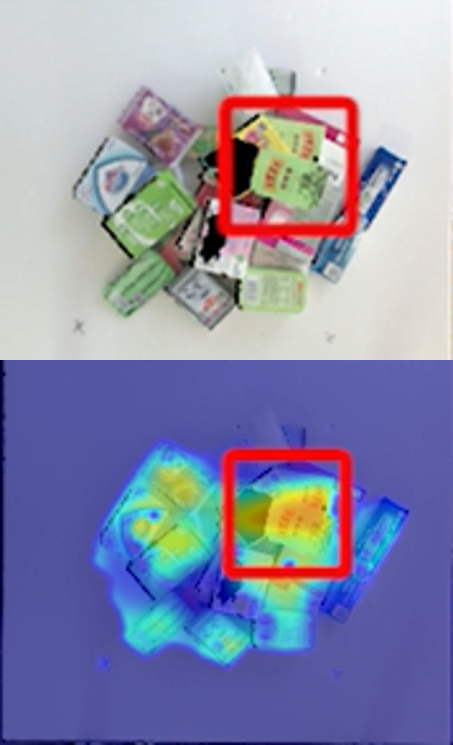}
\end{minipage}
\label{fig:tilte}
}%

\centering
\caption{\textbf{Affordance Map's Failure Case.} Typical failure situations: gathering, covering and tilting. In these situations, the affordance map outputs locations that are not suitable for lifting.}
\end{figure}

In cluttered scenes, the affordance map usually fails in three situations. The first situation is when objects of similar height or color are close to each other (Fig. \ref{fig:gather}). These objects are likely to be regarded as one single object by the affordance ConvNet. In this situation, the junction between adjacent objects will be identified as suitable picking point, which will result in grasp failures. The second situation is when two objects are partially overlapped (Fig. \ref{fig:cover}). The two objects may be treated as one by the affordance ConvNet, and the boundary of the two objects may be identified as suitable picking location. The third situation is that when the pose of the object is over-tilted (Fig. \ref{fig:tilte}). In this case, the picking point indicated by the affordance map may not be suitable for realistic operation, especially when the surface of the object is not smooth enough.
\par

\subsection{Active Exploration}
In order to solve the above problem, the active exploration is introduced into the proposed system. Different from using only one static affordance map, the robot will actively explore and change the environment until a good affordance map is obtained. The deep Q-Network (DQN) is employed to train an agent which indicates actions given the affordance map for the current scene. The network structure (Fig. \ref{fig:DQN_Stru}) is based on the U-Net \cite{unet}, which indicates the pixel-wise action. U-Net is a powerful and lightweight network structure proposed recently for image segmentation, including times of downsampling and upsampling. It demonstrates good performance in yielding pixel-wise semantic information. To minimize the size of the network for speed reason, we adjust this structure to a more tiny one, with one downsampling and upsampling, and resize the RGBD image to a quarter resolution.

\subsubsection{Local patch}
Since our goal is to change the scene according to the current affordance map $I_{aff}$, we don't need to consider the whole scene at each time, which may lead to opposite results. Therefore, we propose a local-patch U-Net structure in the network, which can obtain a better scene with less steps and also minimize the model size for faster computation.\par

Assuming that in the current state, $p_{M}$ is the most promising picking point with the highest confidence score in the affordance map ($p_{M}=argmax\{{I_{aff}\}}$). We crop the input RGBD image around this corresponding pixel with a size of $128 \times 128$ and downsample it to a size of $32 \times 32$ ($32=128/4$) before feeding it into our U-Net based network, which greatly reduces the model size.

\subsubsection{Paralleled tiny U-Net}
The U-Net \cite{unet} is able to indicate pixel-wise actions given image inputs. For each action, it outputs a confidence score on each location. We define 8 specific actions in this work. The robot could push the object from 8 directions with a fixed distance. We use $O_i=i*45^\circ({i=0,\ldots,7)}$ to denote the directions and the push distance is half the size of the local patch. So the whole network contains 8 U-Net modules with the same structure.

The U-Net is trimmed to a tiny one, which down-samples and up-samples for only once. It is good enough for our input and suitable for our scenarios with subpixel-wise operation locations. In this way, the action space of DQN is reduced for a faster learning.

\subsection{The metric of affordance map}
\label{Metric}
Considering the above mentioned failure cases, a novel metric $\Phi$ is designed to calculate the reward for the current affordance map, which is useful for evaluating each action obtained from the DQN. And the next action will be generated accordingly for the robot to change the current scene. The process will be iterated until a good affordance map is obtained.

\subsubsection{Flatness punishment based on Gaussian distribution}

\begin{figure}[h]
\subfigure[affordance of directional distribution]{
	\begin{minipage}[t]{1\linewidth}
		\centering
		\includegraphics[width=3in]{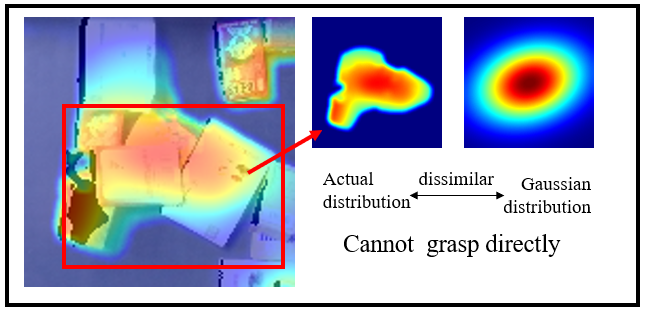}
	\end{minipage}%
}%

\subfigure[affordance of well-distributed distribution]{
	\begin{minipage}[t]{1\linewidth}
		\centering
		\includegraphics[width=3in]{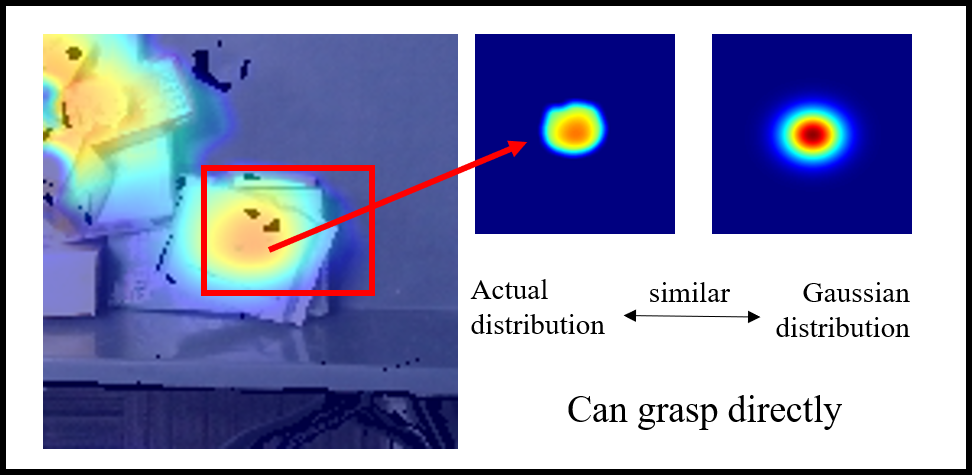}
	\end{minipage}%
}%
\caption{\textbf{The Distribution of Affordance Value.} When the object of the maximum affordance value is tilted or piled up with other objects, the actual distribution of affordance value is directional and different from the Gaussian distribution. When it is not tilted or separate, the actual distribution is well-distributed and similar to the Gaussian distribution.}
\label{Gaussfuction}
\end{figure}

By analyzing the failure cases of the affordance map, it is found that the maximum affordance value appears around the area where there is accumulation or tilting and the distribution of the affordance values around this area is tend to be directional (Fig. \ref{Gaussfuction}). Thus, we extract the connected area near this maximum affordance value and binarize the affordance map. A Gaussian fitting is applied to this area and an estimated affordance value $\hat{s}$ is obtained. With the real affordance value $s$ of this specific area and the maximum affordance value $v_M$, we calculate the standard deviation $\sigma$ of the relative deviation $e$ between $\hat{s}$ and $s$:

\begin{equation}
e_{ij}=\frac{\hat{s}_{ij}-s_{ij}}{v_M}, {\rm{~~~~}}
\sigma=\frac{1}{{m}\cdot{n}}\sqrt{\sum_{i=0}^{m}\sum_{j=0}^{n}{e_{ij}}^{2}}
\end{equation}

When $\sigma$ is small, it indicates that the relative deviation of $\hat{s}$ and $s$ is in a very small range of fluctuations, so the distribution of the affordance value in this connected area is well-distributed. To evaluate the affordance map by $\sigma$, a flat metric $\Phi_{f}$ is introduced as $\Phi_{f}=e^{-\sigma}$.

\subsubsection{Interpeak intervals}

\begin{figure}[h]
\centering
\subfigure[suitable scene for suction]{
\begin{minipage}[t]{0.5\linewidth}
\centering
\includegraphics[width=1.4in]{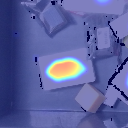}
\end{minipage}%
}%
\subfigure[unsuitable scene for suction]{
\begin{minipage}[t]{0.5\linewidth}
\centering
\includegraphics[width=1.4in]{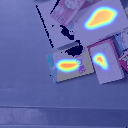}
\end{minipage}
}%
\centering
\caption{\textbf{Leaks of Affordance Map.} When the accumulation of objects occurs near the lift point, there will be often several peaks in the affordance map.}
\label{fig:peak}
\end{figure}

In situation where there are too many objects in the scene, the affordance map often has several peak values. Fig. \ref{fig:peak} shows this difference. We calculate the bounding box of the connected area which is closest to the maximum affordance value and choose the center of this area as one peak location. Besides, we find other peaks by detecting points which have higher values than other points in other small areas. The coordinate of the maximum affordance value point is denoted as $p_M$. The length of the bounding box $l$ and the width $w$ are used to denote the size of the object that will be lifted. Taking $k$ as the number of all the other peaks in the affordance map, $P_{m}=\{p_{0} \cdots p_{k-1}\}$ as the set of other peaks' coordinates, an interval metric $\Phi_{d}$ is defined:

\begin{equation}
a=\frac{w+l}{2}
\end{equation}

\begin{equation}
\Phi_{d}=min\{{\frac{||p_{M}-p_{i}||_2}{a}}, 1\} ({i=0,\ldots,k-1)}
\end{equation}

\subsubsection{Maximum affordance}
We also take the maximum affordance value $v_M$ itself into consideration, which is directly derived from the ConvNet.

\subsubsection{Reward design}
So the final metric $\Phi$ is defined as a weighted sum of the above three metrics:

\begin{equation}
\Phi=\lambda_f*\Phi_f+\lambda_d*\Phi_d+\lambda_v*v_M
\end{equation}
where $\lambda_f+\lambda_d+\lambda_v=1$. So $\Phi \in [0,1]$. And if the metric of current frame $\Phi_i>0.85$, we assume the scene is good enough for grasping and the robot stops changing the environment. Please note that Ref.\cite{active} does not provide the details about the reward design.

Based on the designed metric, the goal of the agent is to maximize the value of this metric. Therefore, if $\Phi_i$ is larger than the metric of last frame $\Phi_{i-1}$ by $\delta$, the reward is 1, otherwise, -1. To reduce the noise interference, we set $\delta=0.01$.

\section{EXPERIMENT}
\label{sec:Experiment}

We test the proposed grasping system by executing a series of experiments both in simulation environment and real-world environment.

\subsection{Experiment of DQN Performance in Simulation Environment}
\begin{figure}[h]

\centering

\includegraphics[width=1\linewidth]{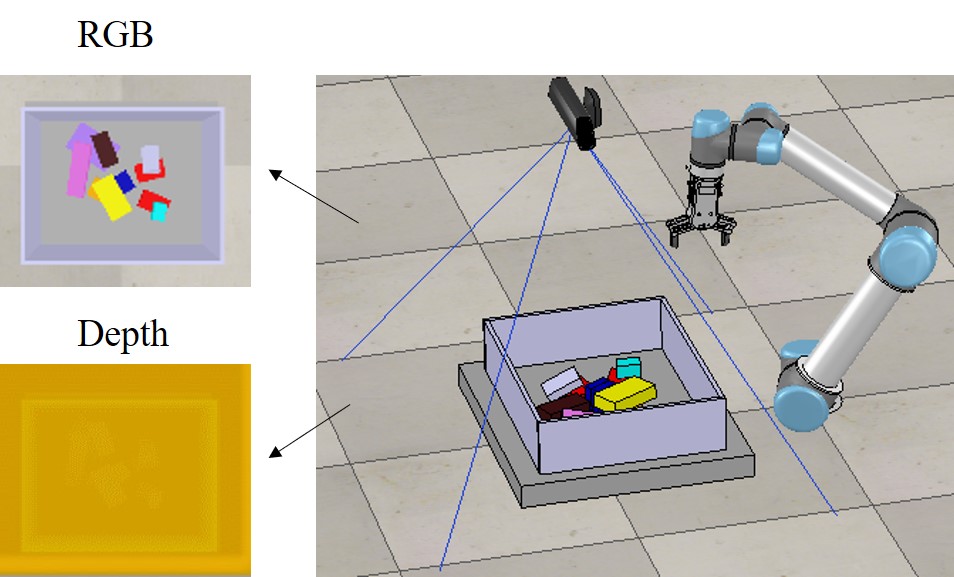}
\caption{\textbf{Simulation Environment}}
\label{fig:simulation setup}
\end{figure}

We choose V-REP\cite{VREP2013} as the simulation environment. The simulation scene (Fig. \ref{fig:simulation setup}) is just the same as that in \cite{active}, where a UR5 manipulator as well as a robotic hand are introduced to implement the process of the active exploration and a Kinect camera is utilized to obtain the visual data. To simulate a cluttered environment, 11 blocks are added into the scene and we manually design several challenging scenes for evaluation. During the training and testing phases, if the $\Phi$ value reaches the threshold, we directly remove the corresponding object.

We compare the proposed model and the random operation model. In the random operation model, instead of pushing the object based on the output of the DQN, a random pushing action is applied. Within 30 continuous operations, if 5 objects are removed, the test finishes and it is called a success. Otherwise, it is called a failure. Three metrics are used to evaluate their performance: 1) average number of operations per test, 2) average increment of metric $\Phi$ per push, and 3) test success rate, which defined as the number of successful tests divided by the number of tests.

In our experiments, we empirically choose the weight parameters as: $\lambda_f=0.75, \lambda_d=0.15, \lambda_v=0.1$. So the metric will be:

\begin{equation}
\Phi=0.75*\Phi_f+0.15*\Phi_d+0.1*v_M
\end{equation}

\subsubsection{Evaluation result}

\begin{figure}[h]
\label{fig:simulation results}
\centering
\includegraphics[width=1\linewidth]{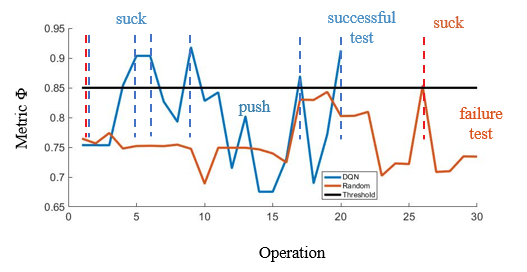}
\caption{\textbf{Metric Over Operations.} Brown line stands for random operations. Blue line stands for DQN-based operation. Black line stands for the metric threshold for object removing. Using DQN-based methods, the metric of affordance map can reach threshold in less steps compared to random operations which proves that DQN has learned intelligent strategies to explore the environment.}
\label{fig:simulation_result}
\end{figure}

The evaluation results in simulation environment are shown in Table \ref{tab:simulation_result} and Fig. \ref{fig:simulation_result}. It can be seen that compared with random operation model, our model trained by DQN can improve the metric $\Phi$ of the affordance map more quickly for a higher grasp success rate and the grasping process is more efficient.

\subsubsection{Training details}
We train our U-Net based DQN model by RMSPropOtimizer, using learning rates decaying from $10^{-3}$ to $2.5 \times 10^{-4}$ and setting the momentum equal to 0.9. Our future discount $\gamma$ is set to be 0.6 for more attention on current epoch. The exploration $\epsilon$ is initialized with 1 and then to 0.2, giving allowance for more attempts on new pushing strategies.

\begin {table}[!h]
\centering
\caption{SIMULATION RESULT OF RANDOM OPERATION AND DQN}
\begin{tabular}{m{2.2cm}<{\centering}|m{1.8cm}<{\centering}|m{1.4cm}<{\centering}|m{1.4cm}<{\centering}}
\hline
Method & Operation times & Metric $\Phi$ increment & Test success rate\\
\hline
Random operation & 23.6 & 0.0216 & 60.0\%\\
\hline
Our model & 20.4 & 0.0219 & 71.4\%\\
    \hline
\end{tabular}
\label{tab:simulation_result}
\end{table}

\subsection{Robotic Experiments}
\subsubsection{Experiment setup}
We choose Microsoft's Kinect V2 camera as the image acquisition tool to get the RGB image and depth image of the scene. The composite robotic hand is mounted on the UR5 manipulator. We select 40 different objects to build different scenes for our robotic hand to grasp.

\subsubsection{Evaluation metric}
The evaluation metrics of real-world experiments are different from those in simulation experiments because the real experiments can be more intuitive. In addition, we find out that when the object of maximum affordance value is unable to be lifted, the robot will repeat this failure operation, as the environment and affordance map are not changed. Therefore, we define a test as failure if the lift fails at the same object for 3 consecutive times, while a test is defined as a success if the 10 objects within a scene are lifted successfully. Based on that, we defined 3 metrics: 1) the average number of objects grasped successfully per test, 2) suction success rate, which is defined as number of objects grasped successfully divided by number of lift operations, 3) test success rate, which is defined as the number of successful tests divided by the number of tests.

\subsubsection{Experiment result}
We test our robotic hand in 20 different scenes with static affordance map and with affordance map optimized by active exploration. The result of every operation is recorded. Experiment results show that after active exploration optimization, the system performs better in suction success rate and test success rate. Compared with lifting only with static affordance map, the active exploration reduces the possibility of repeating failure lifts, making it more robust to the scene. The comparison results are shown in Fig. \ref{fig:results}.

\begin{figure}[h]
\label{fig:results}
\centering
\includegraphics[width=1\linewidth]{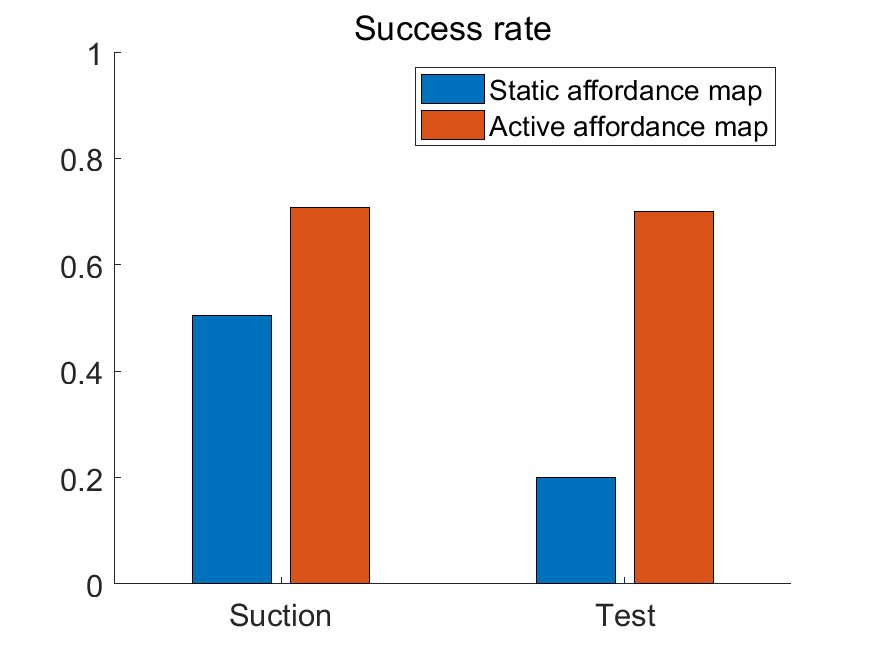}
\caption{Comparison results.}
\label{fig:results}
\end{figure}

\begin{figure}[h]
\centering
\subfigure[typical failure scenes]{
\begin{minipage}[t]{1\linewidth}
\centering
\includegraphics[width=3in]{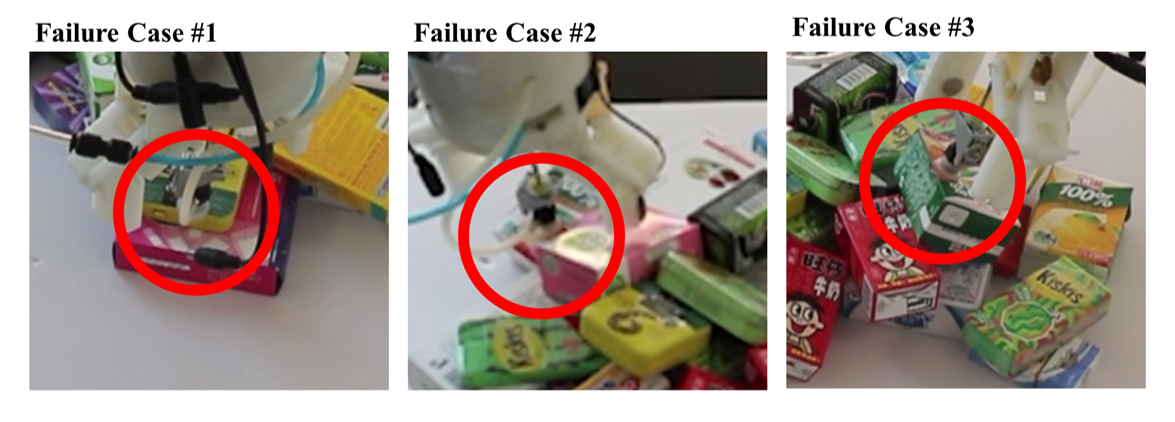}
\end{minipage}%
}%

\subfigure[typical successful scenes]{
\begin{minipage}[t]{1\linewidth}
\centering
\includegraphics[width=3in]{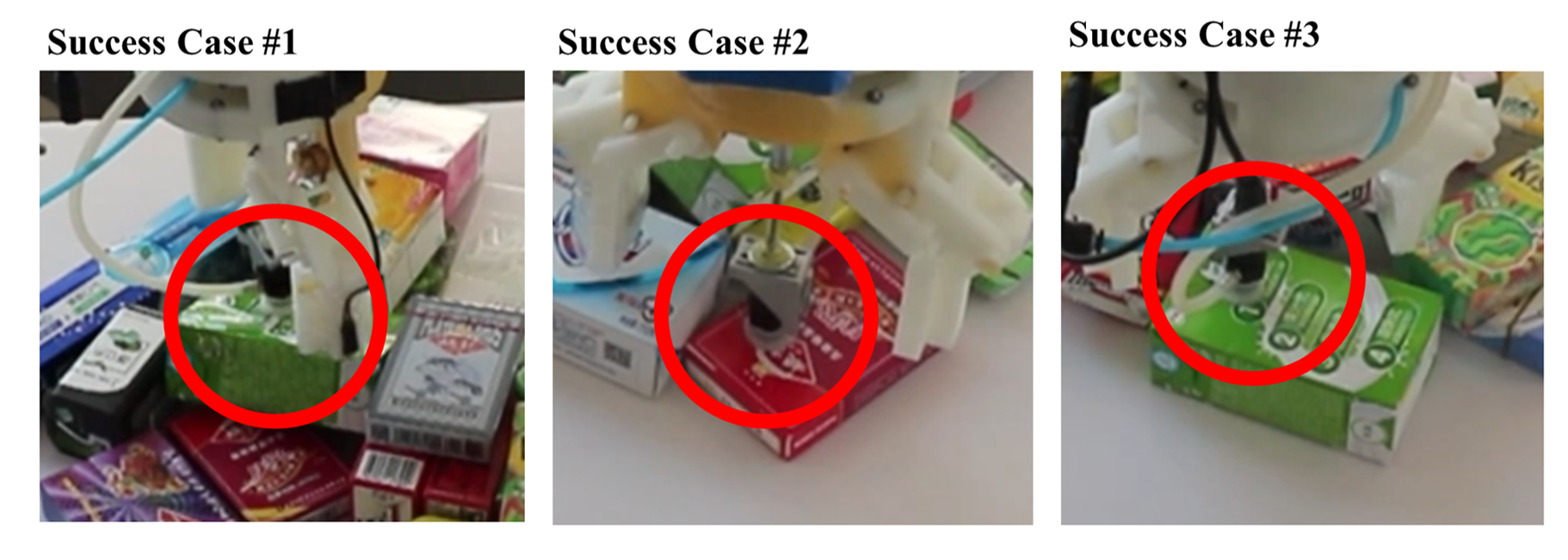}
\end{minipage}%
}%

\centering
\caption{\textbf{Typical Failure Scenes and Successful Scenes.} The failure scenes happen because the affordance map regards several close objects as one single object.}
 \label{Fig: real-word_experiment_static}
\end{figure}

When the system only relies on the static affordance map for grasping, it is likely to fail in cluttered scenes. Fig. \ref{Fig: real-word_experiment_static} shows the result of the grasping experiment using static affordance map.

\begin{figure}[!h]
\subfigure[before disturbance]{
\begin{minipage}[t]{0.5\linewidth}
\centering
\includegraphics[width=1.4in]{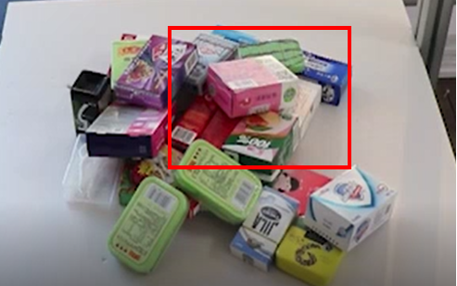}
\end{minipage}%
}%
\subfigure[after disturbance]{
\begin{minipage}[t]{0.5\linewidth}
\centering
\includegraphics[width=1.4in]{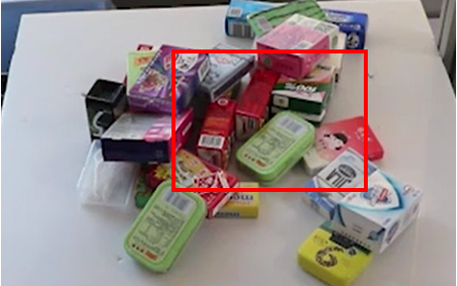}
\end{minipage}%
}%
\caption{\textbf{Exploration Strategy.} Our system changes scene that are not suitable for grasping by pushing in a certain direction with fingers.}
\label{Fig: real-word_experiment_move}
\end{figure}

By using the affordance map optimized by active exploration, it is easier for the system to find grasping point. Therefore, the system can find more reliable grasping point. In Fig. \ref{Fig: real-word_experiment_move}, the robotic hand actively explores the environment to find proper grasping points.

\begin{figure}[!h]
\centering
\subfigure[the object without enough support]{
\begin{minipage}[t]{0.5\linewidth}
\centering
\includegraphics[width=1.4in]{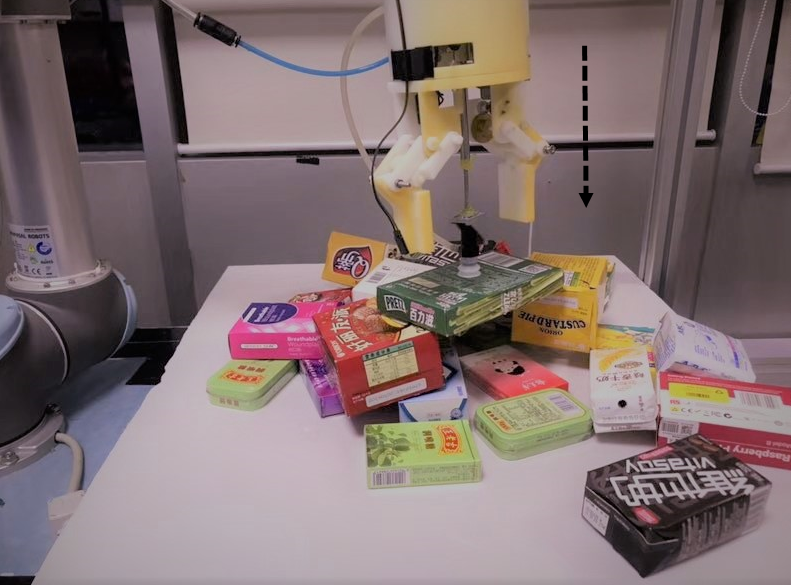}
\end{minipage}%
\label{Figs:no support}
}%
\subfigure[useless push]{
	\begin{minipage}[t]{0.5\linewidth}
		\centering
		\includegraphics[width=1.4in]{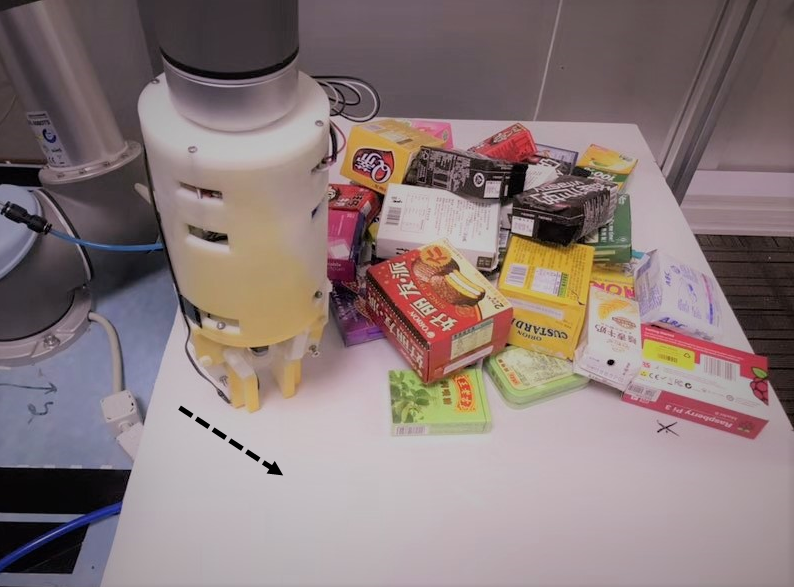}
	\end{minipage}%
\label{Figs:useless push}
}%
\caption{\textbf{Remaining failure cases} with Affordance map optimized by active exploration.}
\label{Fig: failure of actexp}
\end{figure}

However, affordance map optimized by active exploration still has some problems. There are two main problems. The first problem is that the proposed metric $\Phi$ can not distinguish all bad scenes that are not suitable for grasping (Fig. \ref{Figs:no support}). When an object does not have enough support for lifting, our system can not get this information and this kind of objects are difficult for grasping. The second problem is that our DQN sometimes outputs useless pushing action (Fig. \ref{Figs:useless push}) in the area with no objects.

\subsubsection{Result analysis}
In some simple scenes, static affordance map can produce good results by quickly analyzing the whole scene to get the suitable location for grasping. However, in cluttered scenes, it is likely to output incorrect results. Especially, when several objects are very close, the best grasping point in the affordance map will be at the boundary of objects, leading to a failure grasp. What's more, it may always output the same wrong point, making the operation efficiency low. In the proposed model, the affordance map is optimized by active exploration and appropriate operations are generated to push the object away from each other. The situation of the scene is simplified to make the object easy to grasp.

However, the proposed grasping strategy is still not perfect, there exist some useless pushings including pushing in the place without any objects and the system can not recognize all kinds of objects that are not suitable for grasping. In addition, some objects with uneven surface are still difficult to lift with this strategy. As the objects are removed continuously, the scene will become more and more simple. Accordingly, the strategy is gradually inclined to adopt static affordance map directly or with less pushing, so as to ensure the efficiency of grasping.

\section{Conclusion}
In this work, a novel robotic grasping system is proposed, which includes a composite robotic hand which combines a suction cup and a two-finger gripper. At the same time, a DQN based active exploration approach is applied to the system to intelligently grasp object in the cluttered environment. The pushing strategy is used for the robot to actively explore and change the environment until a good affordance map is obtained. It has been demonstrated that it’s more efficient to use the suction cup together with the two-finger gripper for grasping. And the active exploration strategy shows superior performance compared to methods with only static affordance map.\par

\section{Acknowledgements}

This work was supported in part by the National Natural Science Foundation of China under Grants 61703284 and U1613212.


\end{document}